\def\year{2021}\relax
\documentclass[letterpaper]{article} 
\usepackage{aaai21}  
\usepackage{times}  
\usepackage{helvet} 
\usepackage{courier}  
\usepackage[hyphens]{url}  
\usepackage{graphicx} 
\usepackage{natbib}  
\usepackage{caption} 
\usepackage{amsmath}
\usepackage{amssymb}
\usepackage{amsthm}
\usepackage{color}
\usepackage{subcaption}
\usepackage{graphicx}
\usepackage{tabularx}
\usepackage{wrapfig}

\newcommand{\states}{\textit{Q}}
\newcommand{\state}{\textit{q}}

\newcommand{\aut}{\mathcal{T}}

\newtheorem{example}{Example}

\newcommand{\AP}{\mathsf{AP}}

\usepackage{tikz}
\usepackage{pgfplots}
\usepackage[switch]{lineno}  %
\usepackage{algorithm2e}

\DeclareUnicodeCharacter{2212}{−}
\usepgfplotslibrary{groupplots,dateplot}
\pgfplotsset{compat=newest}

\usetikzlibrary{patterns,arrows,arrows.meta,calc,shapes,shadows,decorations.pathmorphing,decorations.pathreplacing,automata,shapes.multipart,positioning,shapes.geometric,fit,circuits,trees,shapes.gates.logic.US,fit}
\tikzstyle{round} = [thick,draw=black,circle]
\usepackage{multirow}
\tikzset{cross/.style={cross out, draw=black, minimum size=2*(#1-\pgflinewidth), inner sep=0pt, outer sep=0pt},
	cross/.default={4pt}}

\title{Temporal-Logic-Based Reward Shaping for Continuing \\ Reinforcement Learning Tasks}
\author {
        Yuqian Jiang\textsuperscript{\rm 1},
        Suda Bharadwaj\textsuperscript{\rm 2},
        Bo Wu\textsuperscript{\rm 2},
        Rishi Shah\textsuperscript{\rm 1,3},
        Ufuk Topcu\textsuperscript{\rm 2},
        Peter Stone\textsuperscript{\rm 1,4}\\
}
\affiliations {
    \textsuperscript{\rm 1} Department of Computer Science, The University of Texas at Austin \\
    \textsuperscript{\rm 2} Department of Aerospace Engineering and Engineering Mechanics, The University of Texas at Austin \\
    \textsuperscript{\rm 3} Amazon \\ 
    \textsuperscript{\rm 4} Sony AI \\
    \{jiangyuqian, suda.b\}@utexas.edu, bowu86@gmail.com, rishihahs@gmail.com, utopcu@utexas.edu, pstone@cs.utexas.edu
}

\newtheorem{theorem}{Theorem}

\newtheorem{remark}{Remark}

\begin{document}
\maketitle
\begin{abstract}
In continuing tasks, average-reward reinforcement learning may be a more appropriate problem formulation than the more common discounted reward formulation.
As usual, learning an optimal policy in this setting typically requires
a large amount of training experiences. Reward shaping is a common approach for
incorporating domain knowledge into reinforcement learning in order to speed
up convergence to an optimal policy. However, to the best of our knowledge, the theoretical properties of reward shaping have thus far only been established in the discounted setting.
This paper presents the first reward shaping framework for average-reward learning and
proves that, under standard assumptions, the optimal policy under the
original reward function can be recovered. In order to avoid the need for manual construction of
the shaping function, we introduce a method for utilizing domain knowledge expressed as a temporal
logic formula. The formula is automatically translated 
to a shaping function that provides additional reward throughout the learning
process. We evaluate the proposed method on three continuing tasks. In all
cases, shaping speeds up the average-reward learning rate without any
reduction in the performance of the learned policy compared to relevant baselines.

\end{abstract}

\section{Introduction}

Reinforcement learning (RL) is a popular method for autonomous agents to learn optimal behavior through repeated interactions with the environment. Most RL algorithms aim to optimize the total \emph{discounted reward} received by the learner. However, in cases involving infinite-horizon or \emph{continuing tasks}, a discount factor can often lead to  undesirable behaviors since the agent sacrifices long-term benefits for short-term gains ~\citep{mahadevan1996average}. Hence, the natural quantity to optimize for many continuing tasks is the \emph{average reward}. However, reinforcement learning with an average reward objective has received very little attention in the literature despite its importance in artificial intelligence~\cite{wan2020learning}. 

Many robotics applications for RL have delayed or sparse rewards, slowing down learning significantly due to long stretches of possibly uninformative exploration~\citep{mahadevan1996average}.  
\emph{Reward shaping}  \citep{225046,mataric1994reward}
is a common way to inject domain knowledge to guide exploration, but requires care so as not to change the underlying RL problem. \cite{randlov98bicycle} added an intuitive signal to the reward of an agent learning to ride a bicycle, but had the unintended effect of causing the agent to ride in circles.
Ng et al.~(\citeyear{ng1999policy}) considered additional rewards that are expressed as the difference of a \emph{potential function} between the agent's current state and its next state and showed that this so-called \emph{potential-based reward shaping} (PBRS) does not affect the optimal policy. Wiewiora et al.~(\citeyear{wiewiora2003principled}) further extended PBRS to include shaping functions of actions as well as states and Devlin \& Kudenko~(\citeyear{devlin2012dynamic}) allow 
time-varying shaping functions. To the best of our knowledge, all of the work in reward shaping  concentrates on the discounted-reward setting, and often for Q-learning in particular. However, the reinforcement learning problem under discounted reward objectives is fundamentally incompatible with continuing learning tasks~\cite{naik2019discounted} and requires a major modification to handle average reward objectives. In this paper, we develop a reward shaping framework for \emph{average-reward} RL.

In general, manually constructing a potential function is difficult since it requires detailed knowledge of the task and the reward structure~\citep{grzes2010online}. To address this challenge, previous works provide methods of learning shaping functions online~\citep{grzes2010online} or offline~\citep{MarthiShaping}. 
In contrast, our focus is on incorporating side-information or domain knowledge into the shaping function. For example, a user may wish to advise a robot tasked with cleaning to spend more time in one room compared to another. While this statement is intuitive for humans, it can be non-trivial to translate such a statement to an appropriate shaping function. We introduce a method to synthesize the shaping function directly from a temporal logic formula representing the domain knowledge. Temporal logic is a structured natural-like language designed to reason about temporal information, which makes it a suitable choice for non-expert users to encode knowledge for continuing tasks, rather than directly interacting with a mathematical function. 
Furthermore, manually constructing a shaping function requires taking into account the relative magnitudes of rewards. In contrast, our proposed method allows for a qualitative input of domain knowledge and does not depend on knowledge of the original reward values and the underlying RL algorithm.

We test the proposed framework in three continuing learning tasks: continual area sweeping~\citep{ahmadi2005continuous, shah2020deep}, control of a cart pole in OpenAI gym~\citep{2016openai}, and motion-planning in a grid world~\citep{mahadevan1996average}. We compare with two benchmarks: the  baseline differential Q-learning  without reward shaping; and shielding, which is a learning method where the learning agent never violates the given temporal logic formula~\citep{alshiekh2018safe}. The proposed framework improves learning performance from the baseline in terms of sample efficiency. 
We also show that when the provided domain knowledge is inaccurate, the proposed method still learns the optimal policy and is faster than the baseline differential Q-learning. At the same time, shielding fails to learn the optimal policy as the learner is forced to strictly follow the inaccurate domain knowledge.  

Our contributions are as follows: 1) we develop a reward shaping method for average-reward RL, 2) we remove the need for potential function engineering by allowing domain knowledge to be specified as a temporal logic formula, 3) we compare the approach to baseline differential Q-learning without reward shaping as well as shielding to illustrate the advantage of our method. 

\section{Related Work}
The goal of this paper is to provide a framework to facilitate high-level interactions with RL algorithms for users without significant RL expertise. Specifically, we are interested in providing a framework to allow users to provide advice to help increase the rate of learning. As such, we focus on providing guarantees of optimality regardless of the quality of advice provided. 

There are several existing areas of research that develop techniques to learn from human knowledge. Learning from demonstration is the problem of learning a policy from examples of a teacher performing a task~\citep{argall2009survey,hussein2017imitation}. Knowledge from a human trainer can be transferred through real-time feedback to improve learning performance~\citep{thomaz2006reinforcement,KCAP09-knox,AAAI18-Warnell}. These techniques usually optimize some combination of inferred human rewards and task rewards. This work focuses on optimizing natural environmental rewards regardless of the quality of knowledge. For continuing tasks, demonstrations or real-time feedback data can be difficult to collect. Our approach takes a temporal-logic specification as input and does not require the human to perform the task or observe the learner during training. 

Since we focus on obtaining advice from users without RL expertise, we use temporal logic to sidestep manual reward engineering. 
In recent years, there has been growing interest in using temporal logic in RL. For example, Icarte et al. proposed using LTL formulas \citep{toro2018teaching} and later reward machines \citep{icarte2018using} to specify reward functions directly. In contrast, our work focuses on using LTL formulas to shape a given reward function from the environment to improve the rate of learning.  
Shaping has also been used to improve reward machines so that the output reward functions are easier to learn for~\cite{camacho2018non, camacho2019ltl}, but it only exploits the values of different reward machine states. Consider a simple specification ``always follow the human.'' The shaping approach introduced by~\citeauthor{camacho2019ltl} (\citeyear{camacho2019ltl}) would only differentiate states based on whether the human is visible now. In contrast, our synthesis of the potential function encourages actions that keep the human visible in the future, assuming knowledge of the transition graph.

Some other approaches use temporal logic formulas as constraints that must be satisfied to guarantee safety during learning \citep{alshiekh2018safe,nilsshield}.  To that end, the actions violating the safety formula are overwritten and never taken by the learning agent. In our setting, these actions are not inherently unsafe but simply "not recommended" based on the provided domain knowledge. Not allowing the agent to take these actions can be overly restrictive, and in cases where the provided domain knowledge is not exactly correct, can lead to sub-optimal policies.



\section{Preliminaries}
In this section, we introduce the definitions and concepts of average-reward reinforcement learning, reward shaping, and temporal logic used throughout the paper.

\subsection{Average-Reward Reinforcement Learning}
A Markov decision process (MDP) $\mathcal{M}=(\mathcal{S},s_I,\mathcal{A},R,P)$ is a tuple consisting of a finite  set $\mathcal{S}$ of states, an initial state $s_I \in \mathcal{S}$, a finite set $\mathcal{A}$ of actions , reward function $R: \mathcal{S}\times \mathcal{A} \times \mathcal{S} \rightarrow \mathbb{R}$, and a probabilistic transition function $P: \mathcal{S} \times \mathcal{A} \times \mathcal{S} \rightarrow [0,1]$ that assigns a probability distribution over successor states given a state and an action. For a stationary policy $\pi:\mathcal{S}\rightarrow\mathcal{A}$, the expected \emph{average reward} is defined as
\begin{equation}\label{equation:average reward function}
    \rho^\pi_\mathcal{M}:=\liminf_{n\rightarrow\infty}\frac{1}{n}\mathbb{E}\left[\sum_{k=0}^{n-1}R(s_k,\pi(s_k),s_{k+1})\right].
\end{equation}

We are interested in reinforcement learning (RL) problems where the objective is to maximize an average reward over time. More specifically,  given an MDP $\mathcal{M}$, where the transition function $P$ and/or reward function $R$ are \emph{unknown} a priori, 
the objective is to learn an optimal policy $\pi^*$ such that $\rho^{\pi^*}_\mathcal{M}\geq\rho^\pi_\mathcal{M}$ for any stationary policy $\pi$.  


Given a policy $\pi$, the $Q$-function $Q_\mathcal{M}^\pi(s,a)$ is the differential value of taking action $a$ in state $s$ and thereafter following $\pi$, and defined as 
\begin{equation}\label{equation:q value function}
    Q_\mathcal{M}^\pi(s,a):=\mathbb{E}\left[\sum_{k=0}^{\infty}R(s_k,a_k,s_{k+1})-\rho_\mathcal{M}^\pi|s_0=s,a_0=a\right].
\end{equation}

The optimal differential $Q$-function $Q_\mathcal{M}^*$ satisfies the Bellman equation \citep{sutton1998introduction}:
\begin{equation}\label{equation:bellman equation}
   \begin{split}
       Q_\mathcal{M}^*(s,a)& = \mathbb{E}[R(s,a,s')+\max_{a'\in\mathcal{A}}(Q_\mathcal{M}^*(s',a'))|P(s,a,s')>0] \\
       &\quad -\rho^*_\mathcal{M},\forall s\in \mathcal{S},a\in\mathcal{A}, 
   \end{split}
\end{equation}
where $\rho^*_\mathcal{M}=\rho^{\pi^*}_\mathcal{M}$, the expected average reward of executing the optimal policy $\pi^*$.
From $Q_\mathcal{M}^*$, the optimal policy $\pi^*$ can be determined by $\pi^*(s) = \arg\max_{a\in\mathcal{A}}(Q^*_\mathcal{M}(s, a))$.

R-learning~\cite{schwartz1993reinforcement,mahadevan1996average} is a classical approach for average-reward RL but lacks theoretical properties. Differential Q-learning~\citep{wan2020learning} is a recent provably convergent approach for learning $Q^*_{\mathcal{M}}$ and $\rho^*_\mathcal{M}$, 
where the standard action-value update step in Q-learning is replaced with one derived from  (\ref{equation:bellman equation}): 
\begin{gather*}
        \delta_t  \xleftarrow{} {r_t + \mathop {\max }\limits_{a_{t+1}} {Q_t}({s_{t + 1}},a_{t + 1}) - {\rho _t}(s_t) - {Q_{t}}({s_t},{a_t})}, \\
        {Q_{t + 1}}({s_t},{a_t}) \xleftarrow{} {Q_{t}}({s_t},{a_t}) + \alpha_t \delta_t, \\
        \rho_{t+1}(s_t) \xleftarrow{} \rho_{t}(s_t) + \eta \alpha_t \delta_t, 
\end{gather*}
where $\delta_t$ is the temporal-difference (TD) error at time step $t$, $\alpha$ is the learning rate and $\eta$ is a positive constant. 


\subsection{Potential-Based Reward Shaping and Potential-Based Advice}
Potential-based reward shaping (PBRS)
augments the reward function $R$ by a shaping reward function $F$, where $F(s,s')=\gamma\Phi(s')-\Phi(s)$,  $\Phi:\mathcal{S}\rightarrow\mathbb{R}$ is a \emph{potential function}, 
and $0<\gamma<1$ is the discount factor. The resulting MDP is $\mathcal{M'}=(\mathcal{S},s_I,\mathcal{A},R',P)$ with $R'=R+F$. It has been proven, for the discounted and episodic settings, that $\mathcal{M}$ and $\mathcal{M'}$ have the same optimal policy~\citep{ng1999policy}. The goal is to learn the optimal policy in $\mathcal{M}'$  with improved sample efficiency compared to learning in $\mathcal{M}$. Potential-based look-ahead advice ~\citep{wiewiora2003potential} further extends  PBRS such that the potential function also depends on actions: $ F(s,a,s',a')=\gamma\Phi(s',a')-\Phi(s,a)$, where $a'$ is chosen by the current policy. 

\subsection{Temporal Logic}



We first define some basic notations. The set of finite sequences $w$ over a set $\Sigma$ is denoted $\Sigma^{*}$, and the set of infinite sequences is denoted $\Sigma^{\omega}$. 

Given an  MDP $\mathcal{M} = (\mathcal{S},s_I,\mathcal{A},R,P)$ and a policy $\pi$, a \emph{path} $\sigma = s_0s_1s_2\dots \in \mathcal{S}^\omega$ is a sequence of states with $s_0=s_I$ and $P(s_i,\pi(s_i),s_{i+1})>0$ for $i\geq 0$. Given a set $\AP$ of atomic propositions (Boolean variables), we introduce the \emph{labelling function} $L: \mathcal{S} \rightarrow 2^{\AP}$. For a path $\sigma = s_0s_1s_2\dots \in \mathcal{S}^\omega$, its corresponding label sequence is $w = w_0w_1\dots \in \left(2^{\AP}\right)^{\omega}$, where $w_i = L(s_i)$. 



A linear temporal logic (LTL) formula $\varphi$ constrains all finite and infinite label sequences $w \in \left(2^{\AP}\right)^{\omega} \cup \left(2^{\AP}\right)^{\ast}$ corresponding to paths in $\mathcal{M}$. For the full semantics of LTL, we refer the reader to~\citep{baier2008principles}. We translate the LTL formula $\varphi$ to an equivalent \emph{deterministic finite automaton} (DFA)~\citep{kupferman2001model}. A DFA is a tuple $\aut^{\varphi} = (\states,\state_I,2^{\AP},\delta,H)$ where $\states$ is a finite set of states, $\state_I \in \states$ is the initial state,  $2^{\AP}$ is a finite input alphabet, $\delta:\states\times2^{\AP} \rightarrow \states$ is a deterministic transition function and $H\subseteq \states$ is a set of \emph{accepting} states. A \emph{run} $\overline{q}$ of $\aut^{\varphi}$ over label sequence $w$ is an infinite sequence $\overline{q} = q_0q_1\dots \in Q^{\omega}$ where $q_0 = q_I$ and $q_{i+1} = \delta(q_i,w_i)$.
The run is \emph{accepted} by $\aut^{\varphi}$ if $q_i \in H$ for all $i \geq 0$.  A label sequence $w$ satisfies $\varphi$ iff the induced run is accepted by $\aut^{\varphi}$. Only a subclass of LTL formulas can be translated to an equivalent DFA with acceptance condition as presented above. We informally refer to such formulas as \emph{safety formulas} and the translated DFA as a safety automaton~\citep{kupferman2001model}.

\begin{figure*}[t]
    \centering
    \subfloat[Gazebo environment for cleaning robot example. \label{fig:cas_gazeboworld}]{\includegraphics[width=0.24\textwidth]{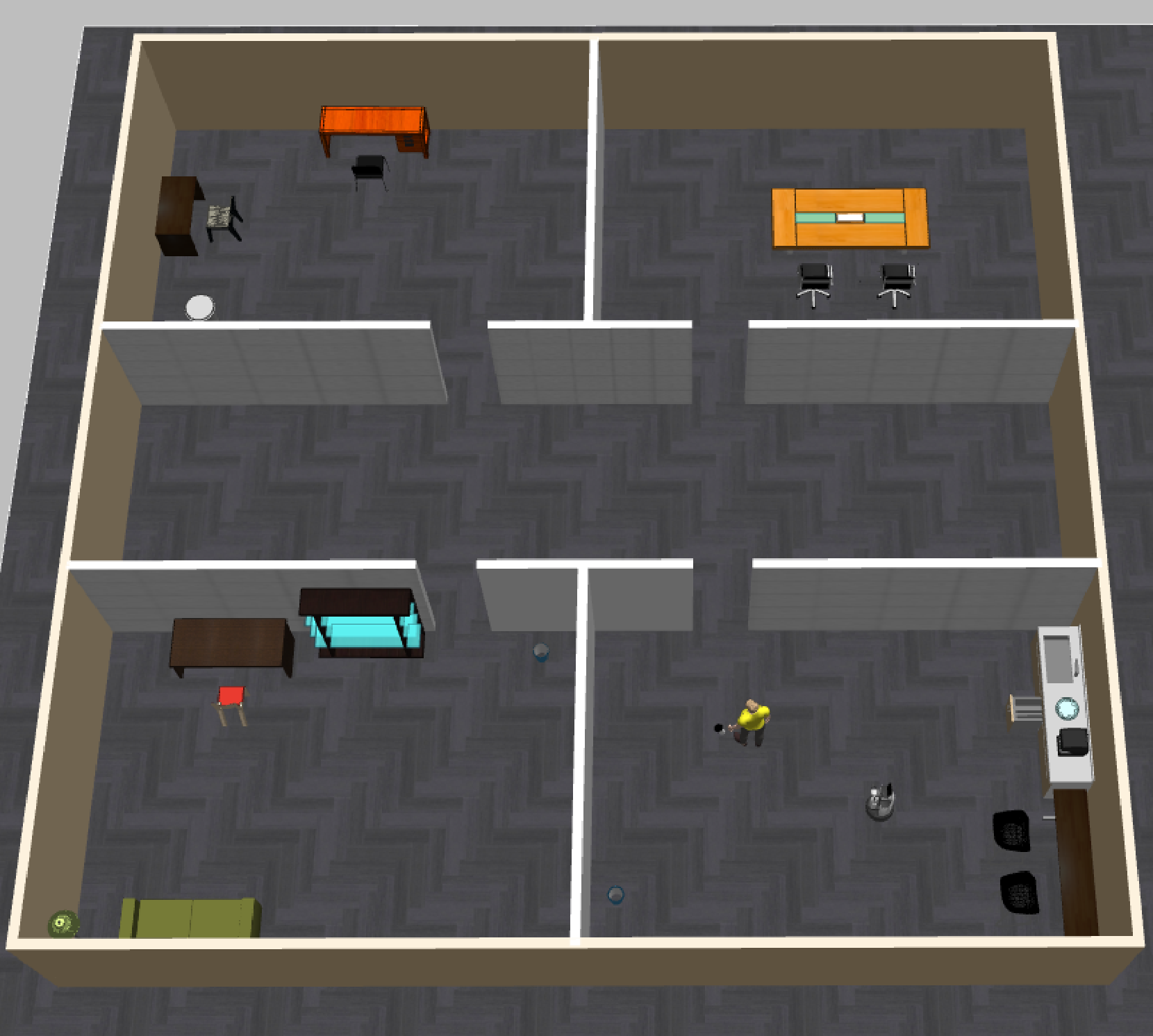}}\hspace{0.25cm}
    \centering
    \subfloat[Grid world representation of (a).  \label{fig:cas_setup}]{ \input{figures/winningregion_gridworld}}
    \subfloat[DFA for $\varphi = \square \; human\_visible$. \label{fig:cas_aut}]{ \begin{tikzpicture}
			\node[state, initial above,accepting]  (q0) {$q_I$};
			\node[state,accepting] [above right= 0.15cm and 2 cm of q0] (q1) {$q_1$};
			\node[state] [below right= 0.15 and 2 cm of q0] (q2) {$q_2$};
			\draw 
			[->] (q0) edge node[below, sloped] {\tiny$\lnot human\_visible$} (q2)
			(q0) edge node[above, sloped] {\tiny $human\_visible$} (q1)
			(q1) edge node[right] {\tiny $\lnot human\_visible$} (q2)
			(q1) edge[loop above] node[left] {\tiny $human\_visible$} (q1)
			(q2) edge[loop below] node[left] {\tiny $\lnot human\_visible$} (q2);
			
			\end{tikzpicture}		}
    \caption{A service robot example where the robot is rewarded for detecting trash that can appear at any state in the gridworld. Both robot and human can move in any of the cardinal directions.}
    \label{fig:cas_setup_full}
\end{figure*}

\begin{example}
Consider a cleaning robot operating in the environment with a human as shown in Figure~\ref{fig:cas_gazeboworld}, which we model as an MDP on a discrete gridworld shown in Figure~\ref{fig:cas_setup}. Trash can appear in any state in the grid with a given probability, and the robot is rewarded when it finds trash.
Suppose we wish to provide the domain knowledge ``the presence of humans tends to increase the probability of trash appearance''. We provide this knowledge in the form of advice as an LTL formula ``Always human visible'' written as  $\square\; human\_visible$. We use this formula to guide the agent during learning to more quickly associate human presence with the appearance of trash. The robot has line-of-sight with a range of 5. We have $\AP = \{human\_visible\}$, and labelling function $L$ that labels all states in visible range as $\{human\_visible\}$ and all other states as $\{\lnot\;human\_visible\}$. The corresponding DFA is shown in Figure~\ref{fig:cas_aut}. 
\end{example}

In the above example, trash appears in states with a higher probability if the state is occupied by a human. However, the given formula $\varphi$ to always keep the human visible, if never violated by the robot,  will likely result in the robot learning suboptimal behavior for the house cleaning task since trash could accumulate in other areas as well. In the next sections, instead of relying on LTL formulas to constrain the RL agent's behavior, we discuss how to use a given LTL formula to \emph{guide} the learning of an optimal policy that may violate the formula.

\section{Reward Shaping for Average-Reward RL}
\label{sec:PBRS_R_learning}



The main contribution of this paper is a novel approach to provide shaping rewards based on an LTL formula to guide a reinforcement learning agent in the average-reward setting while allowing the optimal policy to be learned even if it violates the given formula.
We first define a potential-based shaping function, and prove that the optimal policy can be recovered given any potential function (see supplementary material for the proof).
%

\begin{theorem}
Let $F: \mathcal{S} \times \mathcal{A} \times \mathcal{S} \rightarrow \mathbb{R}$ be a shaping function of the form
\begin{equation}\label{equation:look_ahead}
F(s,a,s')=\Phi\left(s',\arg\max_{a'}(Q^*_\mathcal{M}(s',a'))\right)-\Phi(s,a),
\end{equation}
where $\Phi: \mathcal{S} \times \mathcal{A} \rightarrow \mathbb{R}$ is a real-valued function and $Q^*_\mathcal{M}$ is the optimal average-reward Q-function satisfying (\ref{equation:bellman equation}). Define $\hat{Q}_{\mathcal{M}'}: \mathcal{S} \times \mathcal{A} \rightarrow \mathbb{R}$ as
\begin{equation}\label{equation:Q_hat}
\hat{Q}_{\mathcal{M}'}(s,a):=Q^*_\mathcal{M}(s,a)-\Phi(s,a).
\end{equation}
Then $\hat{Q}_{\mathcal{M}'}$ is the solution to the following modified Bellman equation in $\mathcal{M'}=(\mathcal{S},s_I,A,R',P)$ with $R'=R+F$:
\begin{equation}\label{equation:Q_hat_Bellman}
\hat{Q}_{\mathcal{M}'}(s,a) = \mathbb{E}[R'(s,a,s') +
\hat{Q}_\mathcal{M'}(s',a^*)]-\rho_\mathcal{M'}^{\pi^*},
\end{equation}
where $a^*=\arg\max_{a'}(\hat{Q}_{\mathcal{M}'}(s',a')+\Phi(s',a'))$, and $\rho_{\mathcal{M'}}^{\pi^*}$ is the expected average reward of the optimal policy in $\mathcal{M}$, $\pi^*_\mathcal{M}$, executed in $\mathcal{M'}$. 


\end{theorem}

In MDPs with finite state and action spaces, $\hat{Q}_{\mathcal{M}'}$ is the unique solution to (\ref{equation:Q_hat_Bellman}) because it forms a fully-determined linear system. After learning $\hat{Q}_{\mathcal{M}'}$ with the shaping rewards in $\mathcal{M'}$, it follows from (\ref{equation:Q_hat}) that the optimal policy $\pi^*$ can be recovered as:
$$
\pi^*(s)=\arg\max_{a\in\mathcal{A}}(\hat{Q}_{\mathcal{M'}}(s,a)+\Phi(s,a)).
$$ 
It is important to note that our goal is not to learn the optimal policy of $\mathcal{M'}$. Rather, we learn a value function $\hat{Q}_{\mathcal{M}'}$ from which it is possible to recover the optimal policy of $\mathcal{M}$. Note that $\hat{Q}_{\mathcal{M}'}$ is not the same as $Q_\mathcal{M'}^*$ where $\hat{Q}_{\mathcal{M}'}$ satisfies (\ref{equation:Q_hat_Bellman}) but $Q_\mathcal{M'}^*$ satisfies the following Bellman optimality equation:  
$$
Q_\mathcal{M'}^*(s,a) = \mathbb{E}[R(s,a,s')+\max_{a'\in\mathcal{A}}(Q_\mathcal{M'}^*(s',a'))]-\rho_\mathcal{M'}^{*}.
$$
In practice, the shaping rewards in (\ref{equation:look_ahead}) require knowledge of $Q^*_\mathcal{M}(s',a')$, which can be bootstrapped by  $\hat{Q}_{\mathcal{M}'}(s',a')+\Phi(s',a')$ as its estimate. When shaping is applied to differential Q-learning, action selection is also based on $\hat{Q}_{\mathcal{M}'}(s,a)+\Phi(s,a)$ rather than $\hat{Q}_{\mathcal{M}'}(s,a)$.


\section{Potential Function Synthesis}
We now present an algorithm for constructing a potential function $\Phi$ from a given temporal logic formula $\varphi$. 
Informally, the reward shaping function captures the information in the formula $\varphi$ by \emph{penalizing} the learning agent for visiting the states from which a violation of $\varphi$ \emph{can occur with a non-zero probability}. We illustrate this concept with an example.


\begin{example}
Consider again the example shown in Figure~\ref{fig:cas_setup_full}. In order to not violate the formula $\varphi = \square\;human\_visible$, the robot must ``plan ahead'' to ensure that no matter what the human does in the future, the robot will still not lose visibility of them. Figure~\ref{fig:cas_setup}, shows a case where the current state does not cause a violation of $\varphi$ as the robot can still see the human. However, there is a non-zero probability that the human will move south into a room, and the robot will not be able to maintain visibility and hence will violate $\varphi$.
\end{example}

Informally, we compute the set of state-action pairs from which the probability of violating $\varphi$ is 0. We refer to this set $\mathcal{W}$ as the \emph{almost-sure winning region}. Some knowledge of the dynamics, i.e., the graph structure of $\mathcal{M}$ (not necessarily the transition probabilities), is necessary to ``plan ahead'' and compute $\mathcal{W}$. To this end, we define a \emph{product MDP} $\overline{\mathcal{M}}_{\varphi} = \mathcal{M} \times \aut^{\varphi}$. 
Given an MDP $\mathcal{M} = (\mathcal{S},s_{I},\mathcal{A},R,P) $, a DFA $\aut^{\varphi} = (\states,\state_I,2^{\AP},\delta,H)$, and a labelling function $L$ a product MDP is defined as  $\overline{\mathcal{M}}_{\varphi} = \mathcal{M} \times \aut^{\varphi} = \left(V,v_I,\mathcal{A},\Delta,\overline{H}\right)$ where $V = S \times \states$ is the joint set of states, $v_I = (s_I,\states_I) = (s_I,\delta(q_I,L(s_I)))$ is the initial state, $\Delta: V \times \Sigma \times V \rightarrow [0,1]$ is the probability transition function such that $\Delta((s,q),a,(s',q')) = P(s,a,s')$, if $\delta(q,L(s')) = q'$, and 0, otherwise, and $\overline{H} = (\mathcal{S} \times H) \subseteq V $ is the set of accepting states. 

In order to compute the almost-sure winning region $\mathcal{W} \subseteq \mathcal{S} \times \mathcal{A}$ in the MDP $\mathcal{M}$, we first compute the set $\mathcal{W}^{min}_{0}(\overline{H}) \subseteq V\times \mathcal{A}$ of states and corresponding actions that have a minimum probability of $0$ of leaving $\overline{H}$. This set can be computed using graph-based methods in $\mathcal{O}\left(V\times\mathcal{A}\right)$~\citep{baier2008principles}. An algorithm for computing the set as well as ways to improve scalability for very large state spaces are provided in the supplementary material. We can then extract $\mathcal{W}$ from $\mathcal{W}^{min}_{0}(\overline{H})$ by defining $\mathcal{W} := \{(s,a) \in \mathcal{S}\times\mathcal{A} \mid (s,q,a) \in \mathcal{W}^{min}_{0}(\overline{H}) \}$. 


    


Given $\mathcal{W}$, we construct $\Phi$ as:
     \begin{equation}\label{equation:phi}
         \Phi(s,a) = \begin{cases}
         C & (s,a) \in \mathcal{W} \\
         d(s,a) & (s,a) \notin \mathcal{W} 
         \end{cases}
     \end{equation}
     where $C \in \mathbb{R}$ is an arbitrary constant, and $d: \mathcal{S}\times \mathcal{A} \rightarrow \mathbb{R}$ is any function such that $d(s,a) < C$ for all $s,a$. This construction incentivizes the agent to follow the advice by ensuring that state-action pairs in the almost-sure winning region are assigned higher potential values. 
    \begin{example}
    Continuing the  cleaning robot example, we set $C=1$ and $d(s,a)$ as the negative of the distance between $s$ and the closest state in $\mathcal{W}$. The results are detailed in Section~\ref{sec:personsweeping}. 
    \end{example}
    
    The hyperparameters are formulated generally to allow encoding helpful knowledge about the domain if it is available (such as the Manhattan distance in a grid world), but they can be set arbitrarily as long as $d(s,a)<C$. The results are not sensitive to the precise values, as the shaping function takes the relative difference between the potential values.

\begin{remark}
We note that while the method presented in this section can be applied for any safety formula, the synthesized potential function will not, in general, be Markovian. Since the goal of this paper is to help the agent learn an optimal policy with respect to a Markovian reward function $R(s,a)$, it suffices to consider Markovian potential functions. Restricting $\varphi$ to a subclass of safety formulas called invariant formulas will guarantee $\Phi$ is Markovian ~\citep{baier2008principles}.

\end{remark}

\section{Experiments}
We evaluate our temporal-logic-based reward shaping framework in three average-reward RL benchmarks: continual area sweeping, cart pole, and continuing grid world. In all scenarios, we compare against the standard differential Q-learning approach that we refer to as the \emph{baseline} and the same differential Q-learning approach where actions violating the given temporal logic formula are disallowed using the method introduced by~\citeauthor{alshiekh2018safe} (\citeyear{alshiekh2018safe}). We refer to the latter as \emph{shielding}, and we refer the framework presented in this paper as \emph{shaping}. Furthermore, we study cases when the optimal learned policy must violate the given formula - in such situations, we refer to the provided formula as \emph{imperfect advice}.

We show that 1) our reward shaping framework speeds up baseline average-reward RL, 2) shaping cannot outperform shielding with perfect advice, but 3) when the advice is imperfect, shaping still learns the optimal policy faster than the baseline, whereas shielding fails to learn the optimal policy. All source code is available as supplementary material.



\subsection{Continual Area Sweeping}

The problem of a robot sweeping an area repeatedly and non-uniformly for some task has been formalized as \emph{continual area sweeping}, where the goal is to maximize the average reward per unit time without assuming the distribution of rewards~\cite{shah2020deep}. We run the set of experiments on the environment in Figure~\ref{fig:cas_setup}, where the robot has a maximum speed of 3, i.e. it can move to any cell within a Manhattan distance of 3. We study three cases. First, the robot operates without a human and is given the formula $\square\; kitchen$ where the kitchen is the bottom right room of Figure~\ref{fig:cas_setup}. Second, a human is also present, and the robot is given the formula $\square\; human\_visible$ as described in Examples 1-3. Third, a human is present and there is trash appearing in the corridor, enabling us to study how these approaches handle complex formulae such as conjunctions. The second and the third cases are examples where it is difficult to hand-craft a function, since the winning region is non-trivial, but the advice can be captured with simple and intuitive LTL formulas. In all three scenarios, we compare our framework against the DQN-based deep average-reward RL approach introduced by~\citeauthor{shah2020deep} (\citeyear{shah2020deep}), with their R-learning update replaced by the recent differential Q-learning approach.

\paragraph{Always kitchen}
\begin{figure*}[t!]
\captionsetup[subfigure]{width=11em}
\centering
\subfloat[Trash only appears in the kitchen.\label{fig:cas}]{
\begin{tikzpicture}[scale=.46]
\tikzstyle{every node}=[font=\LARGE]
\definecolor{color0}{rgb}{0.0313725490196078,0.623529411764706,1}
\definecolor{color1}{rgb}{1,0.596078431372549,0.282352941176471}
\definecolor{color2}{rgb}{0.290196078431373,0.525490196078431,0.909803921568627}
\definecolor{color3}{rgb}{1,0.0470588235294118,0.0470588235294118}

\begin{axis}[
legend cell align={left},
legend style={fill opacity=0.8, draw opacity=1, text opacity=1, draw=white!80!black},
legend pos=south east,
tick align=outside,
tick pos=both,
x grid style={white!69.0196078431373!black},
xlabel={Number of Steps},
xmin=500, xmax=10000,
xtick style={color=black},
scaled ticks=false,
tick label style={/pgf/number format/fixed},
y grid style={white!69.0196078431373!black},
ylabel={Average Reward},
ymin=0, ymax=0.8,
ytick style={color=black}
]
\path [draw=color0, fill=color0, opacity=0.3]
(axis cs:500,0.365867863425143)
--(axis cs:500,0.271092136574857)
--(axis cs:1000,0.311355175557671)
--(axis cs:1500,0.359426636467972)
--(axis cs:2000,0.411617194862752)
--(axis cs:2500,0.409236571124078)
--(axis cs:3000,0.443484256603832)
--(axis cs:3500,0.425426938381351)
--(axis cs:4000,0.446358582822536)
--(axis cs:4500,0.422583396940503)
--(axis cs:5000,0.424291967522814)
--(axis cs:5500,0.444973035006899)
--(axis cs:6000,0.452873995916566)
--(axis cs:6500,0.424669083118398)
--(axis cs:7000,0.444849392263841)
--(axis cs:7500,0.449413451300083)
--(axis cs:8000,0.467485774469656)
--(axis cs:8500,0.461947892256591)
--(axis cs:9000,0.470968086482995)
--(axis cs:9500,0.461248398213887)
--(axis cs:10000,0.450262637090193)
--(axis cs:10000,0.569937362909807)
--(axis cs:10000,0.569937362909807)
--(axis cs:9500,0.565071601786113)
--(axis cs:9000,0.567911913517005)
--(axis cs:8500,0.555652107743409)
--(axis cs:8000,0.564874225530344)
--(axis cs:7500,0.563946548699917)
--(axis cs:7000,0.580270607736159)
--(axis cs:6500,0.568090916881602)
--(axis cs:6000,0.561606004083434)
--(axis cs:5500,0.548066964993101)
--(axis cs:5000,0.547948032477186)
--(axis cs:4500,0.533856603059497)
--(axis cs:4000,0.546841417177464)
--(axis cs:3500,0.531093061618649)
--(axis cs:3000,0.537075743396168)
--(axis cs:2500,0.520763428875922)
--(axis cs:2000,0.494182805137248)
--(axis cs:1500,0.433013363532028)
--(axis cs:1000,0.388964824442329)
--(axis cs:500,0.365867863425143)
--cycle;

\path [draw=color1, fill=color1, opacity=0.3]
(axis cs:500,0.463218773900967)
--(axis cs:500,0.396341226099033)
--(axis cs:1000,0.399906562935004)
--(axis cs:1500,0.445865457473064)
--(axis cs:2000,0.477372647003994)
--(axis cs:2500,0.47151159031458)
--(axis cs:3000,0.479751727548689)
--(axis cs:3500,0.472754231150179)
--(axis cs:4000,0.483704767706175)
--(axis cs:4500,0.479866271020535)
--(axis cs:5000,0.482928490528251)
--(axis cs:5500,0.484764866117316)
--(axis cs:6000,0.483399276133835)
--(axis cs:6500,0.484752365850872)
--(axis cs:7000,0.499951624484866)
--(axis cs:7500,0.498252096830925)
--(axis cs:8000,0.501152995720243)
--(axis cs:8500,0.497010129379628)
--(axis cs:9000,0.497770663818748)
--(axis cs:9500,0.496720321803133)
--(axis cs:10000,0.501873753393848)
--(axis cs:10000,0.587046246606152)
--(axis cs:10000,0.587046246606152)
--(axis cs:9500,0.596159678196867)
--(axis cs:9000,0.588909336181252)
--(axis cs:8500,0.584349870620372)
--(axis cs:8000,0.583167004279757)
--(axis cs:7500,0.575747903169075)
--(axis cs:7000,0.577328375515134)
--(axis cs:6500,0.576767634149128)
--(axis cs:6000,0.581760723866165)
--(axis cs:5500,0.572715133882684)
--(axis cs:5000,0.572431509471748)
--(axis cs:4500,0.569733728979465)
--(axis cs:4000,0.569695232293825)
--(axis cs:3500,0.560125768849821)
--(axis cs:3000,0.556448272451311)
--(axis cs:2500,0.56664840968542)
--(axis cs:2000,0.564147352996006)
--(axis cs:1500,0.542974542526936)
--(axis cs:1000,0.470173437064995)
--(axis cs:500,0.463218773900967)
--cycle;

\path [draw=white!65.0980392156863!black, fill=white!65.0980392156863!black, opacity=0.3]
(axis cs:500,0.0948601974152999)
--(axis cs:500,-0.0165801974153)
--(axis cs:1000,-0.0182372365044267)
--(axis cs:1500,-0.0142607911710165)
--(axis cs:2000,0.0113800968204918)
--(axis cs:2500,0.119620242859126)
--(axis cs:3000,0.172478394949465)
--(axis cs:3500,0.278497588220478)
--(axis cs:4000,0.328614398915505)
--(axis cs:4500,0.385185706550257)
--(axis cs:5000,0.419532304319446)
--(axis cs:5500,0.43571744549665)
--(axis cs:6000,0.445741896089562)
--(axis cs:6500,0.446724074811471)
--(axis cs:7000,0.439965764778187)
--(axis cs:7500,0.455411141585395)
--(axis cs:8000,0.46107657986673)
--(axis cs:8500,0.471935082135099)
--(axis cs:9000,0.467741304070571)
--(axis cs:9500,0.468907539818673)
--(axis cs:10000,0.477317169883976)
--(axis cs:10000,0.587562830116024)
--(axis cs:10000,0.587562830116024)
--(axis cs:9500,0.583332460181328)
--(axis cs:9000,0.587058695929428)
--(axis cs:8500,0.576944917864901)
--(axis cs:8000,0.56540342013327)
--(axis cs:7500,0.564468858414605)
--(axis cs:7000,0.574154235221813)
--(axis cs:6500,0.54959592518853)
--(axis cs:6000,0.557338103910438)
--(axis cs:5500,0.54752255450335)
--(axis cs:5000,0.538827695680554)
--(axis cs:4500,0.536094293449743)
--(axis cs:4000,0.550505601084495)
--(axis cs:3500,0.521142411779522)
--(axis cs:3000,0.501441605050534)
--(axis cs:2500,0.437499757140874)
--(axis cs:2000,0.346019903179508)
--(axis cs:1500,0.165220791171017)
--(axis cs:1000,0.113477236504427)
--(axis cs:500,0.0948601974152999)
--cycle;

\addplot [thick, color2]
table {%
500 0.31848
1000 0.35016
1500 0.39622
2000 0.4529
2500 0.465
3000 0.49028
3500 0.47826
4000 0.4966
4500 0.47822
5000 0.48612
5500 0.49652
6000 0.50724
6500 0.49638
7000 0.51256
7500 0.50668
8000 0.51618
8500 0.5088
9000 0.51944
9500 0.51316
10000 0.5101
};
\addlegendentry{Shaping}
\addplot [thick, color3]
table {%
500 0.42978
1000 0.43504
1500 0.49442
2000 0.52076
2500 0.51908
3000 0.5181
3500 0.51644
4000 0.5267
4500 0.5248
5000 0.52768
5500 0.52874
6000 0.53258
6500 0.53076
7000 0.53864
7500 0.537
8000 0.54216
8500 0.54068
9000 0.54334
9500 0.54644
10000 0.54446
};
\addlegendentry{Shielding}
\addplot [thick, white!31.3725490196078!black]
table {%
500 0.03914
1000 0.04762
1500 0.07548
2000 0.1787
2500 0.27856
3000 0.33696
3500 0.39982
4000 0.43956
4500 0.46064
5000 0.47918
5500 0.49162
6000 0.50154
6500 0.49816
7000 0.50706
7500 0.50994
8000 0.51324
8500 0.52444
9000 0.5274
9500 0.52612
10000 0.53244
};
\addlegendentry{Baseline}
\end{axis}

\end{tikzpicture}} 
\centering
\subfloat[Trash can occur outside the kitchen as well.\label{fig:cas_wrong}]{
\begin{tikzpicture}[scale=0.46]
\tikzstyle{every node}=[font=\LARGE]
\definecolor{color0}{rgb}{0.0313725490196078,0.623529411764706,1}
\definecolor{color1}{rgb}{1,0.596078431372549,0.282352941176471}
\definecolor{color2}{rgb}{0.290196078431373,0.525490196078431,0.909803921568627}
\definecolor{color3}{rgb}{1,0.0470588235294118,0.0470588235294118}

\begin{axis}[
legend cell align={left},
legend style={fill opacity=0.8, draw opacity=1, text opacity=1, draw=white!80!black},
tick align=outside,
tick pos=both,
x grid style={white!69.0196078431373!black},
xlabel={Number of Steps},
xmin=500, xmax=10000,
xtick style={color=black},
scaled ticks=false,
tick label style={/pgf/number format/fixed},
y grid style={white!69.0196078431373!black},
ylabel={Average Reward},
ymin=0, ymax=0.8,
ytick style={color=black}
]
\path [draw=color0, fill=color0, opacity=0.3]
(axis cs:500,0.166840911933341)
--(axis cs:500,0.115959088066659)
--(axis cs:1000,0.117573903759224)
--(axis cs:1500,0.156716452748718)
--(axis cs:2000,0.229704942593221)
--(axis cs:2500,0.302118548197401)
--(axis cs:3000,0.284394030183197)
--(axis cs:3500,0.306349805492384)
--(axis cs:4000,0.307568191399499)
--(axis cs:4500,0.310969930706335)
--(axis cs:5000,0.29537013513526)
--(axis cs:5500,0.32528115933541)
--(axis cs:6000,0.33897457646076)
--(axis cs:6500,0.336122048011954)
--(axis cs:7000,0.354734563064559)
--(axis cs:7500,0.372483552319869)
--(axis cs:8000,0.361393563114685)
--(axis cs:8500,0.361419230995041)
--(axis cs:9000,0.37659696731845)
--(axis cs:9500,0.38358215456639)
--(axis cs:10000,0.384210970841738)
--(axis cs:10000,0.506269029158262)
--(axis cs:10000,0.506269029158262)
--(axis cs:9500,0.50109784543361)
--(axis cs:9000,0.495043032681551)
--(axis cs:8500,0.500300769004958)
--(axis cs:8000,0.493886436885315)
--(axis cs:7500,0.483956447680131)
--(axis cs:7000,0.482385436935441)
--(axis cs:6500,0.472837951988046)
--(axis cs:6000,0.45838542353924)
--(axis cs:5500,0.44603884066459)
--(axis cs:5000,0.44114986486474)
--(axis cs:4500,0.430750069293665)
--(axis cs:4000,0.413231808600501)
--(axis cs:3500,0.402490194507616)
--(axis cs:3000,0.391605969816803)
--(axis cs:2500,0.393041451802599)
--(axis cs:2000,0.323335057406779)
--(axis cs:1500,0.224603547251282)
--(axis cs:1000,0.167906096240776)
--(axis cs:500,0.166840911933341)
--cycle;

\path [draw=color1, fill=color1, opacity=0.3]
(axis cs:500,0.177127637622976)
--(axis cs:500,0.131032362377024)
--(axis cs:1000,0.132275124741678)
--(axis cs:1500,0.183161945577894)
--(axis cs:2000,0.253307365642294)
--(axis cs:2500,0.299359425291311)
--(axis cs:3000,0.280732172005526)
--(axis cs:3500,0.295149971456947)
--(axis cs:4000,0.279192618059206)
--(axis cs:4500,0.291900569479776)
--(axis cs:5000,0.286551597360972)
--(axis cs:5500,0.294727707677154)
--(axis cs:6000,0.289133720765657)
--(axis cs:6500,0.298940068300569)
--(axis cs:7000,0.294673845293623)
--(axis cs:7500,0.300111585487361)
--(axis cs:8000,0.295415047131379)
--(axis cs:8500,0.297932945389355)
--(axis cs:9000,0.295725369820824)
--(axis cs:9500,0.295803688663998)
--(axis cs:10000,0.295361253554098)
--(axis cs:10000,0.394278746445902)
--(axis cs:10000,0.394278746445902)
--(axis cs:9500,0.393396311336002)
--(axis cs:9000,0.391514630179176)
--(axis cs:8500,0.404707054610645)
--(axis cs:8000,0.393544952868621)
--(axis cs:7500,0.401088414512638)
--(axis cs:7000,0.393726154706377)
--(axis cs:6500,0.398499931699431)
--(axis cs:6000,0.388706279234343)
--(axis cs:5500,0.394672292322846)
--(axis cs:5000,0.375008402639028)
--(axis cs:4500,0.379699430520224)
--(axis cs:4000,0.368607381940794)
--(axis cs:3500,0.375730028543053)
--(axis cs:3000,0.367147827994473)
--(axis cs:2500,0.382880574708689)
--(axis cs:2000,0.347412634357706)
--(axis cs:1500,0.255118054422106)
--(axis cs:1000,0.175764875258322)
--(axis cs:500,0.177127637622976)
--cycle;

\path [draw=white!65.0980392156863!black, fill=white!65.0980392156863!black, opacity=0.3]
(axis cs:500,0.0913258018144105)
--(axis cs:500,0.0172741981855896)
--(axis cs:1000,0.0156064804683301)
--(axis cs:1500,0.0333218569760004)
--(axis cs:2000,0.121567188940481)
--(axis cs:2500,0.236075571836121)
--(axis cs:3000,0.275969175678791)
--(axis cs:3500,0.2925642070237)
--(axis cs:4000,0.303493204594891)
--(axis cs:4500,0.331698709124748)
--(axis cs:5000,0.344697944000635)
--(axis cs:5500,0.3468776965219)
--(axis cs:6000,0.362377512164268)
--(axis cs:6500,0.365085001914079)
--(axis cs:7000,0.367961526488553)
--(axis cs:7500,0.365091298568881)
--(axis cs:8000,0.377507247984799)
--(axis cs:8500,0.378994615620857)
--(axis cs:9000,0.394337005252664)
--(axis cs:9500,0.385070032348322)
--(axis cs:10000,0.38380302067921)
--(axis cs:10000,0.51383697932079)
--(axis cs:10000,0.51383697932079)
--(axis cs:9500,0.499449967651678)
--(axis cs:9000,0.502662994747336)
--(axis cs:8500,0.501725384379143)
--(axis cs:8000,0.500292752015201)
--(axis cs:7500,0.499468701431119)
--(axis cs:7000,0.491118473511447)
--(axis cs:6500,0.482634998085921)
--(axis cs:6000,0.477982487835732)
--(axis cs:5500,0.4649623034781)
--(axis cs:5000,0.450342055999364)
--(axis cs:4500,0.445421290875252)
--(axis cs:4000,0.428466795405109)
--(axis cs:3500,0.4317557929763)
--(axis cs:3000,0.407350824321209)
--(axis cs:2500,0.368324428163879)
--(axis cs:2000,0.296272811059519)
--(axis cs:1500,0.149638143024)
--(axis cs:1000,0.0984335195316698)
--(axis cs:500,0.0913258018144105)
--cycle;

\addplot [thick, color2]
table {%
500 0.1414
1000 0.14274
1500 0.19066
2000 0.27652
2500 0.34758
3000 0.338
3500 0.35442
4000 0.3604
4500 0.37086
5000 0.36826
5500 0.38566
6000 0.39868
6500 0.40448
7000 0.41856
7500 0.42822
8000 0.42764
8500 0.43086
9000 0.43582
9500 0.44234
10000 0.44524
};
\addlegendentry{Shaping}
\addplot [thick, color3]
table {%
500 0.15408
1000 0.15402
1500 0.21914
2000 0.30036
2500 0.34112
3000 0.32394
3500 0.33544
4000 0.3239
4500 0.3358
5000 0.33078
5500 0.3447
6000 0.33892
6500 0.34872
7000 0.3442
7500 0.3506
8000 0.34448
8500 0.35132
9000 0.34362
9500 0.3446
10000 0.34482
};
\addlegendentry{Shielding}
\addplot [thick, white!31.3725490196078!black]
table {%
500 0.0543
1000 0.05702
1500 0.09148
2000 0.20892
2500 0.3022
3000 0.34166
3500 0.36216
4000 0.36598
4500 0.38856
5000 0.39752
5500 0.40592
6000 0.42018
6500 0.42386
7000 0.42954
7500 0.43228
8000 0.4389
8500 0.44036
9000 0.4485
9500 0.44226
10000 0.44882
};
\addlegendentry{Baseline}
\end{axis}

\end{tikzpicture}}
\centering
\subfloat[Trash only appears where the human has been.\label{fig:cas_person}]{\input{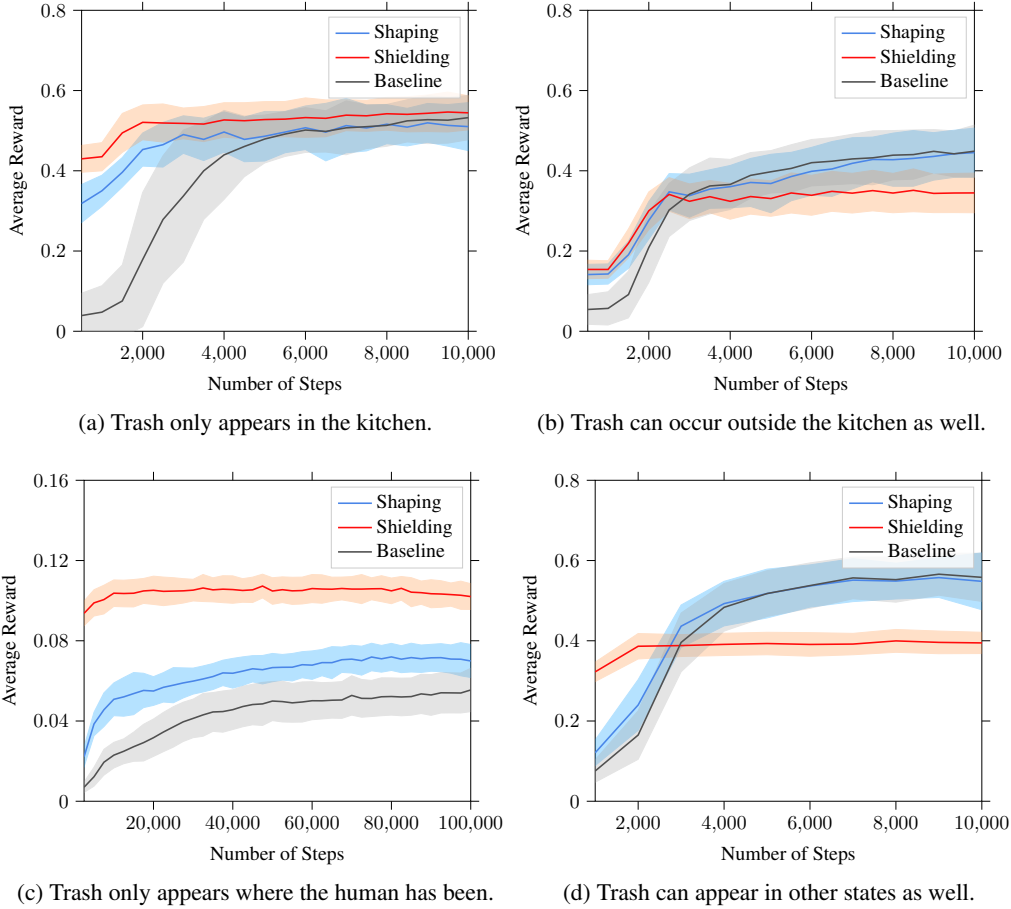}}
\centering
\subfloat[Trash can appear in other states as well.\label{fig:cas_person_wrong}]{
\begin{tikzpicture}[scale=0.46]
\tikzstyle{every node}=[font=\LARGE]
\definecolor{color0}{rgb}{0.0313725490196078,0.623529411764706,1}
\definecolor{color1}{rgb}{1,0.596078431372549,0.282352941176471}
\definecolor{color2}{rgb}{0.290196078431373,0.525490196078431,0.909803921568627}
\definecolor{color3}{rgb}{1,0.0470588235294118,0.0470588235294118}

\begin{axis}[
legend cell align={left},
legend style={fill opacity=0.8, draw opacity=1, text opacity=1,
legend pos=south east,
draw=white!80!black},
tick align=outside,
tick pos=both,
x grid style={white!69.0196078431373!black},
xlabel={Number of steps},
xmin=1000, xmax=10000,
xtick style={color=black},
scaled ticks=false,
tick label style={/pgf/number format/fixed},
y grid style={white!69.0196078431373!black},
ylabel={Average Reward},
ymin=0, ymax=0.8,
ytick style={color=black}
]
\path [draw=color0, fill=color0, opacity=0.3]
(axis cs:1000,0.391858264119109)
--(axis cs:1000,0.344621735880891)
--(axis cs:2000,0.411711224301865)
--(axis cs:3000,0.429604669610945)
--(axis cs:4000,0.460001583999392)
--(axis cs:5000,0.482508980365574)
--(axis cs:6000,0.491111889355274)
--(axis cs:7000,0.500051974729248)
--(axis cs:8000,0.497495592746684)
--(axis cs:9000,0.514241145350454)
--(axis cs:10000,0.516701211817371)
--(axis cs:10000,0.611598788182629)
--(axis cs:10000,0.611598788182629)
--(axis cs:9000,0.606438854649546)
--(axis cs:8000,0.613944407253316)
--(axis cs:7000,0.601988025270752)
--(axis cs:6000,0.584528110644726)
--(axis cs:5000,0.576171019634426)
--(axis cs:4000,0.543838416000608)
--(axis cs:3000,0.518515330389055)
--(axis cs:2000,0.479748775698135)
--(axis cs:1000,0.391858264119109)
--cycle;

\path [draw=color1, fill=color1, opacity=0.3]
(axis cs:1000,0.348875725695757)
--(axis cs:1000,0.300404274304242)
--(axis cs:2000,0.357695903690114)
--(axis cs:3000,0.365832986308542)
--(axis cs:4000,0.359738563377301)
--(axis cs:5000,0.362487618754768)
--(axis cs:6000,0.361830550172286)
--(axis cs:7000,0.361381497261888)
--(axis cs:8000,0.359857204224281)
--(axis cs:9000,0.36026104984491)
--(axis cs:10000,0.365510761587916)
--(axis cs:10000,0.420789238412084)
--(axis cs:10000,0.420789238412084)
--(axis cs:9000,0.42407895015509)
--(axis cs:8000,0.426102795775719)
--(axis cs:7000,0.426298502738112)
--(axis cs:6000,0.421629449827714)
--(axis cs:5000,0.424432381245232)
--(axis cs:4000,0.423361436622699)
--(axis cs:3000,0.425067013691458)
--(axis cs:2000,0.420244096309886)
--(axis cs:1000,0.348875725695757)
--cycle;

\path [draw=white!65.0980392156863!black, fill=white!65.0980392156863!black, opacity=0.3]
(axis cs:1000,0.0968056812068918)
--(axis cs:1000,0.0439943187931082)
--(axis cs:2000,0.105583351612967)
--(axis cs:3000,0.325796697259046)
--(axis cs:4000,0.438584335594161)
--(axis cs:5000,0.473272678718435)
--(axis cs:6000,0.47717531697897)
--(axis cs:7000,0.509993114175886)
--(axis cs:8000,0.508163734165515)
--(axis cs:9000,0.505570829761299)
--(axis cs:10000,0.506553268383115)
--(axis cs:10000,0.630066731616885)
--(axis cs:10000,0.630066731616885)
--(axis cs:9000,0.626689170238701)
--(axis cs:8000,0.614636265834485)
--(axis cs:7000,0.603786885824114)
--(axis cs:6000,0.59650468302103)
--(axis cs:5000,0.570667321281565)
--(axis cs:4000,0.529455664405839)
--(axis cs:3000,0.478703302740954)
--(axis cs:2000,0.222896648387033)
--(axis cs:1000,0.0968056812068918)
--cycle;

\addplot [thick, color2]
table {%
1000 0.36824
2000 0.44573
3000 0.47406
4000 0.50192
5000 0.52934
6000 0.53782
7000 0.55102
8000 0.55572
9000 0.56034
10000 0.56415
};
\addlegendentry{Shaping}
\addplot [thick, color3]
table {%
1000 0.32464
2000 0.38897
3000 0.39545
4000 0.39155
5000 0.39346
6000 0.39173
7000 0.39384
8000 0.39298
9000 0.39217
10000 0.39315
};
\addlegendentry{Shielding}
\addplot [thick, white!31.3725490196078!black]
table {%
1000 0.0704
2000 0.16424
3000 0.40225
4000 0.48402
5000 0.52197
6000 0.53684
7000 0.55689
8000 0.5614
9000 0.56613
10000 0.56831
};
\addlegendentry{Baseline}
\end{axis}

\end{tikzpicture}}
\caption{Learning rate comparison for shaping, shielding, and the baseline in the two continual area sweeping tasks - $\square\; kitchen$ (top row) and $\square\; human\_visible$ (bottom row). In (a) and (b), the average reward of every 500 steps is plotted, averaged by 100 runs.  (c) plots the average reward every 2500 steps, and (d) plots the average reward every 1000 steps, averaged by 100 runs. Only (c) shows the average reward for 100,000 steps (while the rest show 10,000 steps) because the reward is much sparser when trash only appears with human presence. The shaded areas represent one standard deviation from the mean.}
\end{figure*}


In this scenario, the kitchen has the most cleaning needs, and the given formula is to always stay in the kitchen. State-action pair $(s,a) \in \mathcal{W}$ if $L(s) = \{kitchen\}$ and $L(s') = \{kitchen\}$ for all $s'$ such that $P(s,a,s') > 0$. The potential function $\Phi$ is constructed as Equation~\ref{equation:phi}, where $C = 1$ and $d(s,a)$ is the negative of the minimal distance between $s$ and the kitchen and plus $1$ if $a$ gets closer to the kitchen. For the shielding approach, $(s,a)$ is allowed only if $(s,a) \in \mathcal{W}$ or $a$ will decrease the distance to the kitchen from $s$.

Experiments are conducted in the following two scenarios: a) cleaning is required only in the kitchen and nowhere else, and b) cleaning can be required in the kitchen and some states that are randomly selected from the right half of the corridor. Each cell that might require cleaning is assigned a frequency between $1/20$ and $1/10$ so that the robot has to learn an efficient sweeping strategy among those cells. At each time step, there is also a $0.2$ probability that a dirty cell no longer needs cleaning (such as trash gets picked up by people, the wet floor dries as time passes).

Figure~\ref{fig:cas} reports the learning curves when the formula is perfect, i.e., trash only appears in the kitchen. While shielding performs well from the beginning, shaping quickly catches up, and both learn significantly faster than the baseline. Figure~\ref{fig:cas_wrong} reports the results in the second case where there are unexpected rewards outside of the specified region. Shielding gets a head-start by blocking actions away from the region, but it fails to discover the unexpected reward outside the region. Shaping performs better than the baseline at the beginning and converges to a better policy that sweeps in the corridor too.

\paragraph{Always keep human visible}\label{sec:personsweeping}


Consider the scenario described in Examples 1-3. At every time step, there is a $0.2$ probability that the current position of the human needs cleaning. There is also a $0.2$ probability that a dirty cell becomes clean by itself with every step. 
The human moves randomly between the corridor and the top left room and has a speed of 1 cell per step. The state space includes the current location of the human if the human is visible. 
$\Phi$ is constructed as described in Example 3. When the human is invisible to the robot, the winning region is unknown, and $d(s,a)$ is set to $-6$.

Figure~\ref{fig:cas_person} shows the results when the human is the only source of trash appearance. Shielding outperforms shaping and the baseline by forcing the agent to always follow the human. The learning curves of shaping and baseline methods do not converge to the optimal average reward because softmax action selection is used throughout learning. The exploratory actions can lose sight of the human, which is very costly in this task. 

\paragraph{Always keep human visible and always corridor}
Suppose each cell in the corridor is assigned a frequency between $1/20$ and $1/10$ to require cleaning besides the cell that the human occupies at each step. In this case, the best strategy is to always follow the human in the corridor and keep sweeping the corridor when the human goes into the top left room. Shielding cannot use the conjunction of always keeping human visible and always staying in the corridor because there are no allowed actions when the human leaves the corridor. Shaping can easily combine both pieces of advice by adding a potential function based on distance to corridor to the potential function in Example 3. Figure~\ref{fig:cas_person_wrong} shows that the shielding method fails to learn the optimal policy if it is forced to strictly follow the human and misses the reward of staying the corridor; the shaping method is able to leverage both specifications even though they sometimes contradict each other, and outperforms both shielding and the baseline.


\subsection{Cart Pole}


\begin{figure*}[t]
\centering
\subfloat[Continuing cart pole setup.\label{fig:cartpole_setup}]{\includegraphics[width=0.25\textwidth]{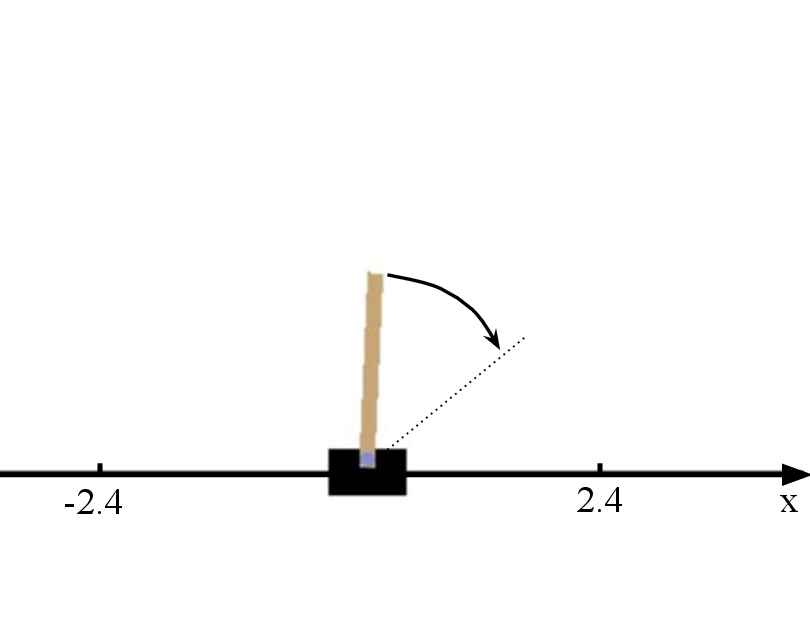}}\hspace{0.1cm}
\centering
\subfloat[Perfect knowledge.\label{fig:cartpole}]{\input{figures/cartpole.tex}}
\centering
\subfloat[Unexpected cart range.\label{fig:cartpole_wrong}]{\input{figures/cartpole_wrong.tex}}
\caption{Learning rate comparison between shaping, shielding, and the baseline in the continuing cart pole task}
\end{figure*}

Cart pole is a widely studied classic control problem in the reinforcement learning literature. The agent is a cart with a pole attached to a revolute joint. Situated on a horizontal track, the agent must keep its pole balanced upright by applying a force to the right or the left. Each state is composed of the position $x$, the linear velocity $\dot{x}$, the pole angle $\theta$, and the angular velocity $\dot{\theta}$. At each step, the agent receives reward of 1 if $x \in [-2.4, 2.4]$ and $\theta \in [-\pi/15, \pi/15]$. Otherwise, the episode terminates.
We modify the cart pole environment in OpenAI Gym~\cite{2016openai} to a continuing task by removing the termination conditions. 
As shown in Figure~\ref{fig:cartpole_setup}, we allow the cart to move anywhere on the $x$-axis and the pole to be at any angle, and restrict the max linear velocity by $|\dot{x}| \leq 1$. Other aspects of the motion model are kept the same, and reward is given under the same conditions. Therefore, the agent has to learn a policy that keeps the cart position in the scoring range (i.e., $x \in [-2.4,2.4]$) and swings the pole up when it falls down. 
 
We compare our shaping framework with the same baseline deep differential Q-learning and shielding. The specification is to keep the cart position in the scoring range. We test the methods in the case of not knowing the full dynamics model, and approximate the winning region by predicting the next x position as $x_{t+1}=x_t + \dot{x_t} \Delta t$ where $\Delta t$ is the time between steps. The shield blocks the action in the direction of $x_{t+1}$ if $x_{t+1}$ is not in the scoring range. $\Phi(s,a)$ is constructed with $C=1$ and $d(s,a)$ as the minimal distance between the scoring range and $x_{t+1}$ (and multiplied by a constant\footnote{It is intuitive to penalize the agent for going out of the scoring range by scaling $d(s,a)$. We tried scaling factors of 1, 10, and 100, and saw the most improvements with 100. Overall, no fine tuning was required for the hyperparameters $C$ and $d(s,a)$.}).


Figure~\ref{fig:cartpole} plots the learning performance when the accurate scoring range $[-2.4, 2,4]$ is given to shaping and shielding. Figure~\ref{fig:cartpole_wrong} shows the results when an inaccurate scoring range $[-2,2]$ is used. Because the cart can leave the scoring range and never reset, the baseline gets extremely sparse reward and fails to solve the problem in a reasonable amount of time in most trials. Shaping and shielding are both able to significantly improve the learning performance, but when the knowledge is inaccurate, shielding over-restricts action choices and achieves worse average reward than shaping. The results also show that shaping is robust to only knowing approximate dynamics.


\subsection{Grid World}


\begin{figure*}[h!]
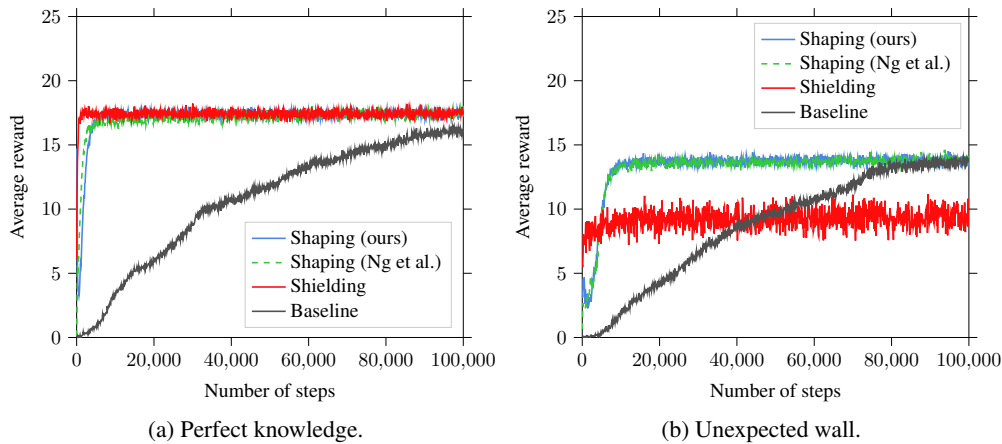

    \centering
    \subfloat[Grid world setup.\label{fig:gridworld_setup}]{\raisebox{.3\height}
        {\begin{tikzpicture}[scale=1]
        \draw[step=0.5cm,color=gray] (-1.5,-1.5) grid (1.5,1.5);
    \filldraw[fill=blue,draw=black] (-1.25,1.25) circle (0.2cm);
    \filldraw[fill=red,draw=black] (-0.5,0.5) rectangle (1.5,0.0);
    \filldraw[fill=green,draw=black] (1.0,-1.0) rectangle (1.5,-1.5);
    \node at (1.30,-1.25) {\tiny{\textbf{G}}};
\end{tikzpicture}

}}\hspace{0.2cm}
    \centering
    \subfloat[Perfect knowledge.]{\input{figures/gridworld_simple}}
    \centering
    \subfloat[Unexpected wall.]{\input{figures/gridworld_wall}}
    
    \caption{Grid world experiments comparing the performance of temporal-logic-based reward shaping (ours), hand-crafted reward shaping (Ng et al.), shielding, and baseline. The average reward of every 100 steps is plotted, averaged by 100 runs. }
    \label{fig:gridworld}
\end{figure*}

We further evaluate our method in a continuing grid world environment. The purpose of this experiment is to include a common benchmark that has been studied in the context of both average-reward RL~\cite{mahadevan1996average} and reward shaping~\cite{ng1999policy}, and showcase that our temporal-logic-based approach performs as well as the hand-crafted shaping function. Figure~\ref{fig:gridworld_setup} shows the 6*6 grid with a +100 reward in the green cell. The agent can move to a neighboring cell in one of the four cardinal directions deterministically at every step, and gets ``transported'' to a random cell when it reaches the green cell and when it fails to reach the green cell after 100 steps. 

The following four methods are compared: our temporal-logic-based reward shaping, hand-crafted reward shaping, shielding, and the baseline. Shielding blocks actions up and left and only allows actions down and right. For temporal-logic-based shaping, the winning region consists of all states, but only with the actions down and right. The potential function $\Phi$ is constructed with $C=1$ and $d(s,a)=-1$. The hand-crafted potential function is introduced by Ng et al. (\citeyear{ng1999policy}), where $\Phi(s)$ is defined as the negative of the distance from $s$ to the green cell. 
The correctness of the advice is varied by adding the red wall from (2, 2) to (2, 5) as shown in Figure~\ref{fig:gridworld_setup}. When the agent is directly above the wall, it cannot reach the green cell by only moving down or right. 

Figure~\ref{fig:gridworld} compares the learning performance of each method in the two conditions. The observations are consistent with the results reported in the other benchmarks. Shielding outperforms other methods when the advice is perfect, i.e., there is no wall, and fails to reach the maximum average reward when the wall is added. Our temporal-logic-based reward shaping method performs similarly to the hand-crafted shaping function while allowing the reward shaping to be automatically constructed. Both shaping methods still outperform the baseline differential Q-learning when the advice is imperfect.

\section{Conclusion}
We provide a reward shaping framework to construct shaping functions from a temporal logic specification to improve performance of average-reward RL. We prove and empirically show that our shaping framework speeds up the learning rate, and allows the optimal policy to be learned when the provided advice is imperfect. 

Since this is the first work to extend reward shaping to the average-reward setting (to the best of our knowledge), we focus on providing theoretical guarantees and intuitive examples rather than evaluations in a large suite of benchmarks. Given that average-reward RL remains under-studied and thus lacks a wide range of benchmarks, this paper also contributes the Continual Area Sweeping task and our continuing version of Cart Pole as new possible benchmarks.

This paper focuses on the (not uncommon) case when the user has some knowledge of the dynamics. In robotics applications, knowledge of the underlying graph is a reasonable assumption as the high-level dynamic capabilities of the robot are generally known even if there is low-level variation which is captured by the detailed (unknown) transition probabilities. The proposed approach is also robust to the user only knowing approximate dynamics, similar to imperfect advice. One direction for future research is extending the work to apply when the dynamics or models are fully unknown.

Another interesting direction for future work that this paper opens up is studying \emph{adversarial} advice - i.e., formulas provided to actively hinder the learning process. We are also interested in expanding the learning problem to include both non-Markovian reward structures and shaping functions. 

\clearpage
\newpage 

\section*{Acknowledgements}
This work was partially supported by ONR N00014-20-1-2115, ARO W911NF-20-1-0140. 

This work has partially taken place in the Learning Agents Research
Group (LARG) at UT Austin.  LARG research is supported in part by NSF
(CPS-1739964, IIS-1724157, NRI-1925082), ONR (N00014-18-2243), FLI
(RFP2-000), ARO (W911NF-19-2-0333), DARPA, Lockheed Martin, GM, and
Bosch.  Peter Stone serves as the Executive Director of Sony AI
America and receives financial compensation for this work.  The terms
of this arrangement have been reviewed and approved by the University
of Texas at Austin in accordance with its policy on objectivity in
research.

We would like to thank all reviewers of this paper for their constructive feedback.

\section*{Ethics Statement}
The guiding philosophy behind the work in this paper is that the barrier to entry for reinforcement learning is too high in terms of domain expertise and data availability. We believe this is particularly exacerbated in robotics applications as robotics has become integrated in various research fields, many of which involve no computer scientists and data collection is costly. RL as a tool has huge potential to unlock the use of robotics for research in many fields, but often requires a significant amount of tuning, training, and expertise to take advantage of properly. Our goal is to make this paper be the first of a series of papers aimed at facilitating high-level interaction with RL algorithms without requiring significant RL expertise and improving the sampling efficiency.  

Since the work at its current stage is still theoretical, there is little direct negative societal consequences. However, this work, as is the case for most robotics-related research, contributes to the increased adoption of robots and autonomous systems in society. In particular, RL allows autonomous systems to achieve complex tasks that they were previously not capable of. In this paper, we use cleaning robots as the motivating case study throughout the paper. Autonomous systems being able to reliably complete complex tasks has the potential to displace the service industries, surveillance, and search and rescue, where the tasks are repetitive or even dangerous.

While we acknowledge the potential job loss that can occur from maturation of such technology, we believe the unlocking the potential of RL to non-experts can lead to the acceleration of research in many fields. 

\newpage
\section*{Appendix}
\section*{Proof of Theorem 1}
We first restate the following equations to be used in the proof.

For a stationary policy $\pi:\mathcal{S}\rightarrow\mathcal{A}$, the expected \emph{average reward} is defined as
\begin{equation}\label{equation:average reward function}
    \rho^\pi_\mathcal{M}:=\liminf_{n\rightarrow\infty}\frac{1}{n}\mathbb{E}\left[\sum_{k=0}^{n-1}R(s_k,\pi(s_k),s_{k+1})\right].
\end{equation}
The optimal $Q$-function $Q_\mathcal{M}^*$ for average reward satisfies the Bellman equation \citep{sutton1998introduction}:
\begin{equation}\label{equation:bellman equation}
   \begin{split}
       Q_\mathcal{M}^*(s,a)& = \mathbb{E}\left[R(s,a,s')+\max_{a'\in\mathcal{A}}(Q_\mathcal{M}^*(s',a'))\right] -\rho^*_\mathcal{M},
   \end{split}
\end{equation}
$\forall s\in \mathcal{S},a\in\mathcal{A}. $

\begin{theorem}
Let $F: \mathcal{S} \times \mathcal{A} \times \mathcal{S} \rightarrow \mathbb{R}$ be a shaping function of the form
\begin{equation}\label{equation:look_ahead}
F(s,a,s')=\Phi\left(s',\arg\max_{a'}(Q^*_\mathcal{M}(s',a'))\right)-\Phi(s,a),
\end{equation}
where $\Phi: \mathcal{S} \times \mathcal{A} \rightarrow \mathbb{R}$ is a real-valued function and $Q^*_\mathcal{M}$ is the optimal average-reward Q-function satisfying (\ref{equation:bellman equation}). Define $\hat{Q}_{\mathcal{M}'}: \mathcal{S} \times \mathcal{A} \rightarrow \mathbb{R}$ as
\begin{equation}\label{equation:Q_hat}
\hat{Q}_{\mathcal{M}'}(s,a):=Q^*_\mathcal{M}(s,a)-\Phi(s,a).
\end{equation}
Then $\hat{Q}_{\mathcal{M}'}$ is the solution to the following modified Bellman equation in $\mathcal{M'}=(\mathcal{S},s_I,A,R',P)$ with $R'=R+F$:
\begin{equation}\label{equation:Q_hat_Bellman}
\hat{Q}_{\mathcal{M}'}(s,a) = \mathbb{E}[R'(s,a,s') +
\hat{Q}_\mathcal{M'}(s',a^*)]-\rho_\mathcal{M'}^{\pi^*},
\end{equation}
where $a^*=\arg\max_{a'}(\hat{Q}_{\mathcal{M}'}(s',a')+\Phi(s',a'))$, and $\rho_{\mathcal{M'}}^{\pi^*}$ is the expected average reward of the optimal policy in $\mathcal{M}$, $\pi^*_\mathcal{M}$, executed in $\mathcal{M'}$. 


\end{theorem}

\begin{proof}
We first show that $\rho_{\mathcal{M}}^{\pi^*}=\rho_{\mathcal{M'}}^{\pi^*}$, which means the expected average reward of $\pi^*$ does not change with the additional rewards from $F$.

From \eqref{equation:average reward function}, we know that
\begin{equation*}
\begin{split}
\rho_{\mathcal{M'}}^{\pi^*}&=\liminf_{n\rightarrow\infty}\frac{1}{n}\mathbb{E}\left[\sum_{k=0}^{n-1}R(s_k,a_k,s_{k+1})+F(s_k, a_k, s_{k+1})\right]\\
              &=\liminf_{n\rightarrow\infty}\frac{1}{n}\mathbb{E}\left[\sum_{k=0}^{n-1}R(s_k,a_k,s_{k+1}) \right. \\
              &\qquad \qquad \qquad \qquad \left. + \Phi(s_{k+1}, \arg\max_{a'}(Q^*_\mathcal{M}(s_{k+1},a'))) \right. \\
              &\qquad \qquad \qquad \qquad \left.- \Phi(s_{k}, a_{k})\vphantom{\sum_{k=0}^{n-1}}\right].
\end{split}
\end{equation*}

Since $a_{k+1} = \pi^*(s_{k+1}) = \arg\max_{a'}(Q^*_\mathcal{M}(s_{k+1},a'))$, we have
\begin{equation}\label{equation:average_reward}
\begin{split}
\rho_{\mathcal{M'}}^{\pi^*} &=\liminf_{n\rightarrow\infty}\frac{1}{n}\mathbb{E}\left[\sum_{k=0}^{n-1}R(s_k,a_k,s_{k+1})+\Phi(s_{k+1}, a_{k+1}) \right. \\
              &\qquad \qquad \qquad \qquad \left. - \Phi(s_{k}, a_{k}) \vphantom{\sum_{k=0}^{n-1}}\right] \\
              &=\liminf_{n\rightarrow\infty}\frac{1}{n}\mathbb{E}\left[\sum_{k=0}^{n-1}R(s_k,a_k,s_{k+1}) \right. \\
              &\qquad \qquad \qquad \left. +\Phi(s_{n},a_{n})-\Phi(s_{0},a_{0})\vphantom{\sum_{k=0}^{n-1}}\right]\\
              &=\liminf_{n\rightarrow\infty}\frac{1}{n}\mathbb{E}\left[\sum_{k=0}^{n-1}R(s_k,a_k,s_{k+1})\right]\\
              &=\rho_\mathcal{M}^{\pi^*}.
\end{split}
\end{equation}

Subtracting $\Phi(s,a)$ from both sides of the Bellman optimality equation (\ref{equation:bellman equation}) gives us
\begin{equation}\label{equation:Q*}
\begin{split}
& \quad Q_\mathcal{M}^*(s,a)-\Phi(s,a) \\ =& \quad \mathbb{E}\left[R(s,a,s')-\Phi(s,a)+\max_{a'\in\mathcal{A}}(Q_\mathcal{M}^*(s',a'))\right]-\rho_\mathcal{M}^{\pi^*}.
\end{split}
\end{equation}

Substituting $\hat{Q}_{\mathcal{M}'}(s,a):=Q^*_\mathcal{M}(s,a)-\Phi(s,a)$ and $a^*=\arg\max_{a'}(\hat{Q}_{\mathcal{M}'}(s',a')+\Phi(s',a'))$ in (\ref{equation:Q*}), we get
\begin{equation*}
\begin{split}
\hat{Q}_{\mathcal{M}'}(s,a) 
& = \mathbb{E} \left[\vphantom{\max_{a'\in\mathcal{A}}} R(s,a,s')-\Phi(s,a) \right. \\
&\qquad \left. +\max_{a'\in\mathcal{A}}(\hat{Q}_{\mathcal{M}'}(s',a')+\Phi(s',a')) \right]-\rho_\mathcal{M}^{\pi^*} \\
& = \mathbb{E} \left[\vphantom{\hat{Q}_\mathcal{M'}} R(s,a,s')+\Phi(s',a^*)-\Phi(s,a) \right. \\
&\qquad \left. + \hat{Q}_\mathcal{M'}(s',a^*)\right]-\rho_\mathcal{M}^{\pi^*}\\
&\stackrel{(\ref{equation:average_reward})}{=} \mathbb{E} \left[\vphantom{\hat{Q}_\mathcal{M'}} R(s,a,s')+F(s,a,s') \right. \\
&\qquad \left. + \hat{Q}_\mathcal{M'}(s',a^*)\right]-\rho_\mathcal{M'}^{\pi^*}\\
& = \mathbb{E}\left[R'(s,a,s') + \hat{Q}_\mathcal{M'}(s',a^*)\right]-\rho_\mathcal{M'}^{\pi^*}.
\end{split}
\end{equation*}

\end{proof}

\newpage
\section*{Algorithm to compute almost-sure winning region}

In the following, we present the algorithm to compute the almost-sure winning region. Informally, the goal of the algorithm is to compute a set of state-action pairs from which there is a \emph{minimum probability 0} of violating the safety formula $\varphi$. We repeat some definitions here for self-completion. 

Given an MDP $\mathcal{M} = (\mathcal{S},s_{I},\mathcal{A},R,P) $, a DFA $\aut^{\varphi} = (\states,\state_I,2^{\AP},\delta,H)$, and a labelling function $L$ a product MDP is defined as  $\overline{\mathcal{M}}_{\varphi} = \mathcal{M} \times \aut^{\varphi} = \left(V,v_I,\mathcal{A},\Delta,\overline{H}\right)$ where $V = S \times \states$ is the joint set of states, $v_I = (s_I,\states_I) = (s_I,\delta(q_I,L(s_I)))$ is the initial state, $\Delta: V \times \Sigma \times V \rightarrow [0,1]$ is the probability transition function such that $\Delta((s,q),a,(s',q')) = P(s,a,s')$, if $\delta(q,L(s')) = q'$, and 0, otherwise, and $\overline{H} = (\mathcal{S} \times H) \subseteq V $ is the set of accepting states.

In order to compute the almost-sure winning region $\mathcal{W} \subseteq \mathcal{S} \times \mathcal{A}$ in the MDP $\mathcal{M}$, we first compute the set $\mathcal{W}^{min}_{0}(\overline{H}) \subseteq V\times \mathcal{A}$ of states and corresponding actions that have a minimum probability of $0$ of reaching $V \setminus \overline{H}$. This set can be computed using graph-based methods in $\mathcal{O}\left(V\times\mathcal{A}\right)$~\citep{baier2008principles} and the algorithm is given below. We then extract $\mathcal{W}$ from $\mathcal{W}^{min}_{0}(\overline{H})$ by defining $\mathcal{W} := \{(s,a) \in \mathcal{S}\times\mathcal{A} \mid (s,q,a) \in \mathcal{W}^{min}_{0}(\overline{H}) \}$. 

We introduce some additional notation. Let $\text{Act}(v) \subseteq \mathcal{A}$ be the set of all available actions $a \in \mathcal{A}$ available at $v$.

\begin{algorithm}
\KwResult{Set $\mathcal{W}^{min}_{0}(\overline{H})$ of state-action pairs with minimum probability 0 of reaching $V\setminus \overline{H}$}
initialize $\overline{\mathcal{W}} = (V\setminus\overline{H}) \times \mathcal{A}$, $\hat{H} = V\setminus \overline{H}$;\\
\While{True}{
\ForAll{$v \in  \overline{H}$}{
    \ForAll{$a \in \text{Act}(v) $}{
        \If{$\;\exists\; v' \in \hat{H} \text{ such that }  \Delta(v,a,v') > 0$}{$ \overline{\mathcal{W}}= \overline{\mathcal{W}} \cup \{(v,a)\}$}
    }
    \If{ $(v,a) \in \overline{\mathcal W} \;\;\forall\, a \in \text{Act}(v)$}{$ \hat{H}= \hat{H} \cup \{v\}$}
}
$\mathcal{W}^{min}_{0}(\overline{H}) = (V\times \mathcal{A}) \setminus \overline{W}$\\
\Return $\mathcal{W}^{min}_{0}(\overline{H})$ 

}
\caption{Almost-sure winning set construction}
\label{alg:winset}
\end{algorithm}

We note that the algorithm above is a variation of the algorithm to compute the set of states with $\text{Pr}^{\text{min}}(s \models \lozenge B)$ found in~\citep{baier2008principles}. Informally, the algorithm computes the set of states from which there is a minimum 0 probability of \emph{eventually reaching} set B. 

\citeauthor{baier2008principles}(\citeyear{baier2008principles}) proved that the worst case complexity of the approach is quadratic in $|\mathcal{S}||\mathcal{A}|$. One way to handle very large state space is to compute the winning region on decomposed or abstract state spaces. In fact, the method by \citeauthor{alshiekh2018safe}(\citeyear{alshiekh2018safe}) computes a “safety-relevant quotient MDP” that is directly applicable to constructing a smaller MDP for this algorithm.


\bibliographystyle{aaai21}
\bibliography{ref}

\end{document}